# Image contrast enhancement using fuzzy logic

Samrudh. K
Dept. of Electrical & Electronics
Siddaganga Institute of Technology
Tumkur, India
samrudhkumar@hotmail.com

Sandeep Joshi
Dept. of Electronics & Communication
Siddaganga Institute of Technology
Tumkur, India
Sandeepjoshi1910@gmail.com

*Abstract*— **Image enhancement is a method of improving the quality of an image and contrast is a major aspect. Traditional methods of contrast enhancement like histogram equalization results in over/under enhancement of the image especially a lower resolution one. This paper aims at developing a new Fuzzy Inference System to enhance the contrast of the low resolution images overcoming the shortcomings of the traditional methods. Results obtained using both the approaches are compared.**

## I. INTRODUCTION

### A. Image enhancement

Image enhancement is simply a technique which improves the quality of the image, increases the perceptibility of the image which is quintessential in the fields such as medical imaging, surveillance, remote sensing etc. Further this acts as a preprocessing for applications like segmentation, recognition etc.

### B. Histogram

Histogram is important in image processing as it acts as a graphical representation of the tonal distribution in a digital image. It is a graph showing the number of pixels in an image at each different intensity value found in that image.

### C. Fuzzy Logic

Human brain is capable of making excellent decisions using imprecise & incomplete sensory information provided by the perceptive organs. Fuzzy theory provides a systematic calculus to deal with such information linguistically and perform numerical computations using linguistic labels in the form of membership functions. Fuzzy inference system (FIS) when selected properly can effectively model the human expertise in the specific application.

This paper proposes an FIS which can enhance the contrast of low resolution images effectively, these results are then compared with the traditional histogram equalization and various evaluation metrics are tabulated.

## II. HISTOGRAM EQUALIZATION

Histogram is a technique in image processing to spread the histogram of the image evenly over the entire discrete quantization levels. Usually it is applied to low contrast images whose histogram is concentrated around very few discrete levels. This is an effective method but often results in over or under enhancement. In this paper, contrast enhancement is done using histogram equalization and compared with the proposed fuzzy enhancement algorithm. The algorithm for histogram equalization is given as follows.

a) Find the histogram of the image.
b) Find the running sum of the histogram values
c) Normalize the sum by dividing each pixel by total number of pixels or resolution of the image
d) Multiply the obtained values by maximum gray value and round it.
e) Map the obtained values using one to one correspondence.

## III. PROPOSED FUZZY CONTRAST ENHANCEMENT ALGORITHM

We have developed an FIS which takes discrete intensity value of a pixel which is fuzzified using input membership functions, then based on the IF – THEN rules, input is mapped to the output. Finally a defuzzified value is obtained using the output membership functions.

### A. Input membership functions

The system uses 7 input membership functions to make the system more accurate, also Gaussian as well as trap membership functions are used which represents sets of pixel intensity values as linguistic variables like dark, gray, bright etc. shown in fig1. These membership functions were chosen after analyzing many images and range of pixel values for these linguistic variables were set.

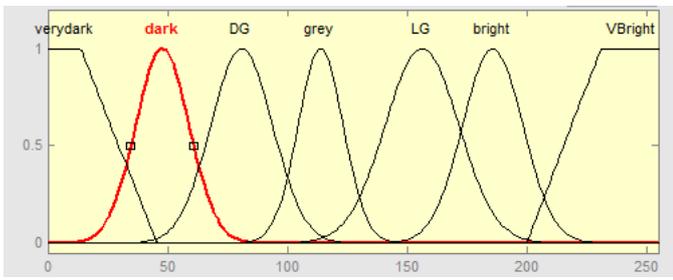
**Fig.1 Input membership functions**

A typical Gaussian curve is defined by

$$G(x,c,\sigma) = e^{-\frac{1}{2}\left(\frac{x-c}{\sigma}\right)^2} \quad (1)$$

Where c represents the center of the membership function and $\sigma$ determines the width.

*B. Output membership functions*

The output membership functions define the amount by which the pixel intensity should be increased or decreased based on the rules defined in the knowledge base. The crisp intensity modification value is obtained from the output membership functions through the process of defuzzification. The output membership functions are shown in fig.2. The basic intuitive idea of contrast enhancement being; if a pixel is dark, make it darker and if a pixel is bright, make it brighter. Based on this idea, intuitive fuzzy inference system is designed.

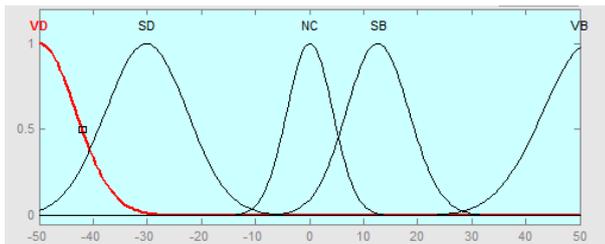
**Fig.2 Output membership functions**

*C. Fuzzy inference system*

The FIS was designed using fuzzy logic toolbox in Matlab. The algorithm uses mamdani FIS. The FIS contains a knowledge base formulated by an expert, this contains IF-THEN rules. These rules map the fuzzy inputs to fuzzy outputs and takes place through compositional rule of intuition. The block diagram of a typical FIS is shown in Fig.3.

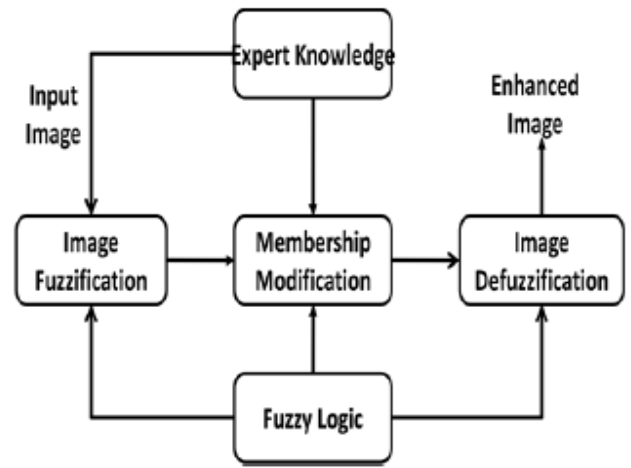
**Fig.3 A typical FIS**

The following are the rules in the proposed algorithm.
1) IF input is Very Dark(VD) THEN output is Slightly Dark(SD)
2) IF input is Dark Gray(DG) THEN output is Slightly Dark(SD)
3) IF input is Gray(G) THEN output is Slightly Dark(SD)
4) IF input is Bright(B) THEN output is Slightly Bright(SB)
5) IF input is Dark(D) THEN output is Very Dark(VD)
6) IF input is Very Bright(VB) THEN output is No Change(NC)
7) IF input is Light Gray(LG) THEN output is Slightly Dark(SD)

## IV. RESULTS

In order to determine the efficacy of the system, we have compared the results obtained with histogram equalization. Various low contrast images were taken and fed into the Matlab and the obtained results are as follows.

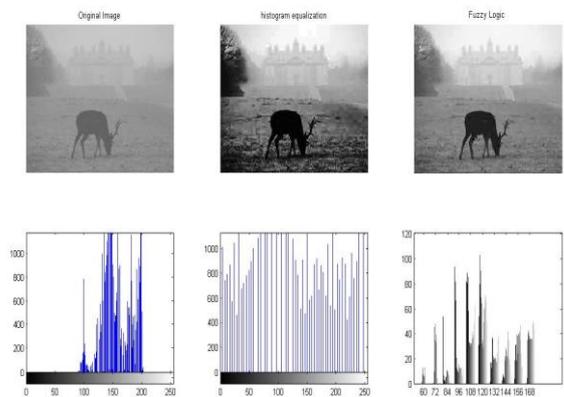
**Fig.4 Deer**

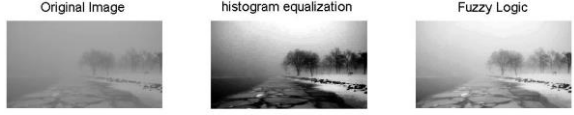

**Fig.5 Lake**

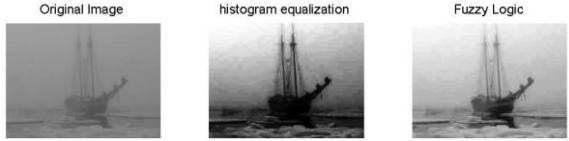

**Fig.6 Ship**

The low contrast images, histogram equalized image and fuzzy logic processed image and their respective histograms are presented in fig.4, 5 and 6.

## V. EVALUATION METRICS

Various evaluation metrics were calculated for images enhanced using histogram equalization and fuzzy logic. The evaluation metrics used are described as follows.

1. Peak Signal to Noise Ratio(PSNR)

PSNR is the ratio of maximum possible power of a signal to power of the corrupting noise. PSNR is an approximation to human perception of reconstruction quality. Higher the PSNR, better is the enhancement, but this may lead to over/under enhancement. Initially, mean square error is calculated and then PSNR is found. Their respective formulae is given in equation 2 and equation 3 respectively.

$$MSE = \frac{1}{mn}\sum_{i=1}^{m}\sum_{j=1}^{n}(x_e - x_0)^2 \quad (2)$$

$$PSNR = 10\log_{10}\left(\frac{L_{max}^2}{MSE}\right) \quad (3)$$

2. Measure of Luminance Index(MLI)

Measure of Luminance index is used as a measure of intensity. This metric is a similarity based approach and it is defined as the ratio between the mean of enhanced image Xe and mean of original image Xo. For a high quality of enhanced image MLI must be high. It is given by equation 4.

$$MLI = \frac{MI(X_e)}{MI(X_o)} \quad (4)$$

Where MI indicates mean of evaluated and original image respectively.

The obtained results are tabulated as follows. Table 1 presents the evaluation of the image "Deer", Table 2 presents the evaluation of image "Lake" and Table 3 presents the evaluation of image "Ship" respectively.

Table 1: Evaluation of image Deer

| Image | Deer | | |
|---|---|---|---|
| Evaluation Metric | Original | Hist. Eqn | Fuzzy |
| Mean | 156.2849 | 127.4542 | 126.3528 |
| MLI | | 0.8155 | 0.8085 |
| MSE | | 5.63E+03 | 895.9924 |
| PSNR | | 10.6241 dB | 15.2879 dB |

Table 2: Evaluation of image Lake

| Image | Lake | | |
|---|---|---|---|
| Evaluation Metric | Original | Hist. Eqn | Fuzzy |
| Mean | 146.1035 | 127.4191 | 116.1912 |
| MLI | | 0.8721 | 0.7953 |
| MSE | | 5.5930e+03 | 894.8265 |
| PSNR | | 10.6543 dB | 13.5893 dB |

Table 3: Evaluation of image Ship

| Image | Ship | | |
|---|---|---|---|
| Evaluation Metric | Original | Hist. Eqn | Fuzzy |
| Mean | 135.4567 | 127.6632 | 105.6157 |
| MLI | | 0.9425 | 0.7797 |
| MSE | | 5.6859e+03 | 890.6227 |
| PSNR | | 10.5828 | 12.8485 |